\documentclass[10pt,twocolumn,letterpaper]{article}

\usepackage[pagenumbers]{cvpr} 

\usepackage{graphicx}
\usepackage{amsmath}
\usepackage{amssymb}
\usepackage{booktabs}

\usepackage{cuted}
\usepackage{capt-of}

\usepackage[pagebackref,breaklinks,colorlinks]{hyperref}

\usepackage[capitalize]{cleveref}
\crefname{section}{Sec.}{Secs.}
\Crefname{section}{Section}{Sections}
\Crefname{table}{Table}{Tables}
\crefname{table}{Tab.}{Tabs.}

\def\cvprPaperID{*****} 
\def\confName{CVPR}
\def\confYear{2022}

\begin{document}

\title{Generating Texture for 3D Human Avatar from a Single Image using Sampling and Refinement Networks}

\author{
Sihun Cha \quad Kwanggyoon Seo \quad Amirsaman Ashtari \quad Junyong Noh\\
\\
Visual Media Lab, KAIST\\
}
\maketitle

\begin{strip}
    \centering
    \includegraphics[width=0.95\textwidth]{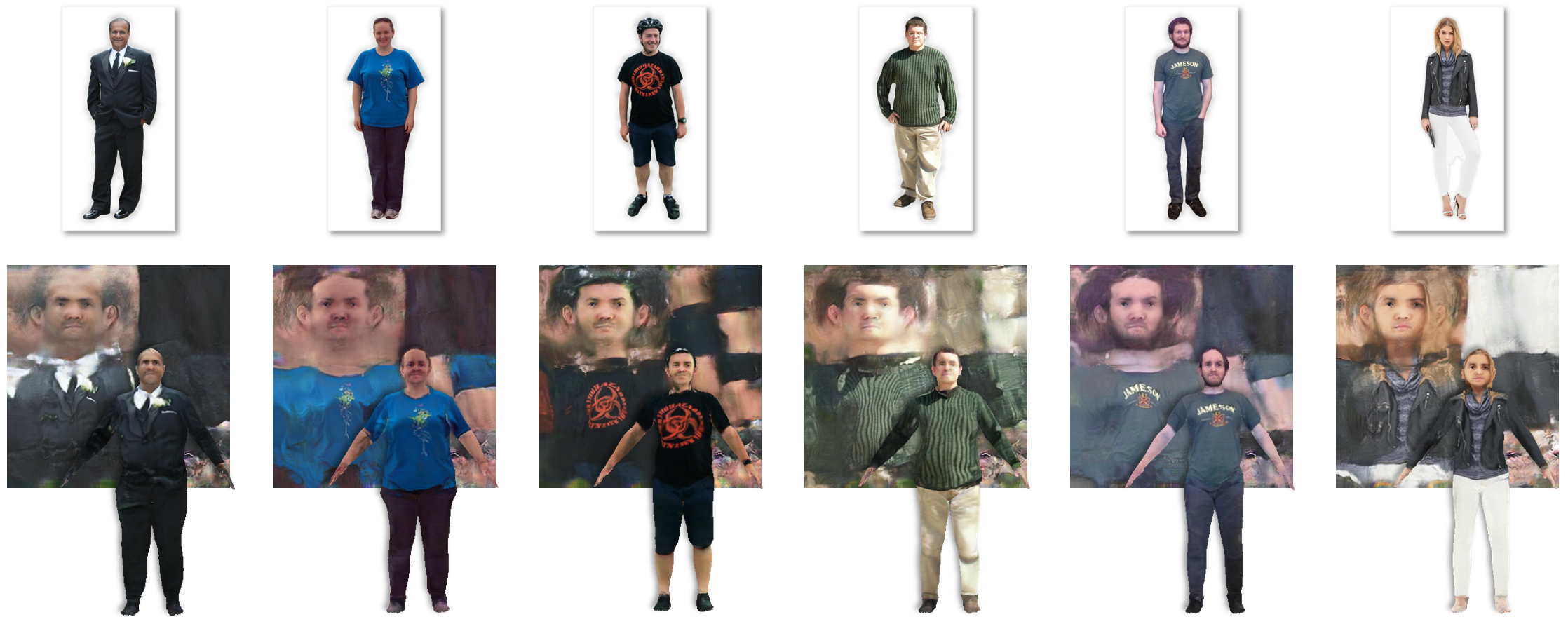}
    \captionof{figure}{
    Given a single image, our method generates a texture map by synthesizing textural patterns in the invisible regions of the source image as well as aligning the texture to the surface of the geometry. The top row shows the source human images and the bottom row shows the rendered images of the 3D human avatars with generated texture maps. The 3D human mesh was obtained using Tex2Shape~\cite{alldieck2019tex2shape} and the images were sampled from SHHQ\cite{fu2022stylehuman} dataset.
    \label{fig:teaser}}
\end{strip}

\begin{abstract}
   There has been significant progress in generating an animatable 3D human avatar from a single image. However, recovering texture for the 3D human avatar from a single image has been relatively less addressed. Because the generated 3D human avatar reveals the occluded texture of the given image as it moves, it is critical to synthesize the occluded texture pattern that is unseen from the source image. To generate a plausible texture map for 3D human avatars, the occluded texture pattern needs to be synthesized with respect to the visible texture from the given image. Moreover, the generated texture should align with the surface of the target 3D mesh. In this paper, we propose a texture synthesis method for a 3D human avatar that incorporates geometry information. The proposed method consists of two convolutional networks for the sampling and refining process. The sampler network fills in the occluded regions of the source image and aligns the texture with the surface of the target 3D mesh using the geometry information. The sampled texture is further refined and adjusted by the refiner network. To maintain the clear details in the given image, both sampled and refined texture is blended to produce the final texture map. To effectively guide the sampler network to achieve its goal, we designed a curriculum learning scheme that starts from a simple sampling task and gradually progresses to the task where the alignment needs to be considered. We conducted experiments to show that our method outperforms previous methods qualitatively and quantitatively.

\end{abstract}

\section{Introduction}

The demand for animatable 3D human avatars is increasing in various VR/AR applications such as virtual try-on, metaverse, and games. To create an animatable 3D human avatar, it is essential to produce a 3D model that resembles the shape and appearance of the source human appearance. Furthermore, the 3D model should be rigged for animation. These processes often require manual work from artists or rely on a special capture system such as multi-view camera sets or 3D scanners. To alleviate these conditions, numerous methods have been proposed to reconstruct a 3D human avatar~\cite{kolotouros2019convolutional,alldieck2019tex2shape,li20193d,lazova2019360,xu20213d,natsume2019siclope,saito2019pifu,saito2020pifuhd,he2020geo,huang2020arch,he2021arch++,alldieck2022photorealistic} from a single image. In contrast to the reported successes in reconstructing body shapes and poses, restoring the occluded texture for 3D human avatars has been relatively less studied.

Generating a texture map for a 3D human avatar from a single image is challenging due to the following two reasons. First, only portions of the texture information are available from the source image. This is caused by various poses and shapes of the human body and the diverse camera positions. Second, the generated human texture map needs to be semantically aligned with the surface of the target 3D human mesh. As the texture coordinates correspond to the surface geometry of a 3D human mesh, misalignment of the texture can produce a distorted human appearance in a rendered image.

\begin{figure}[t]
\centering
  \includegraphics[width=\linewidth]{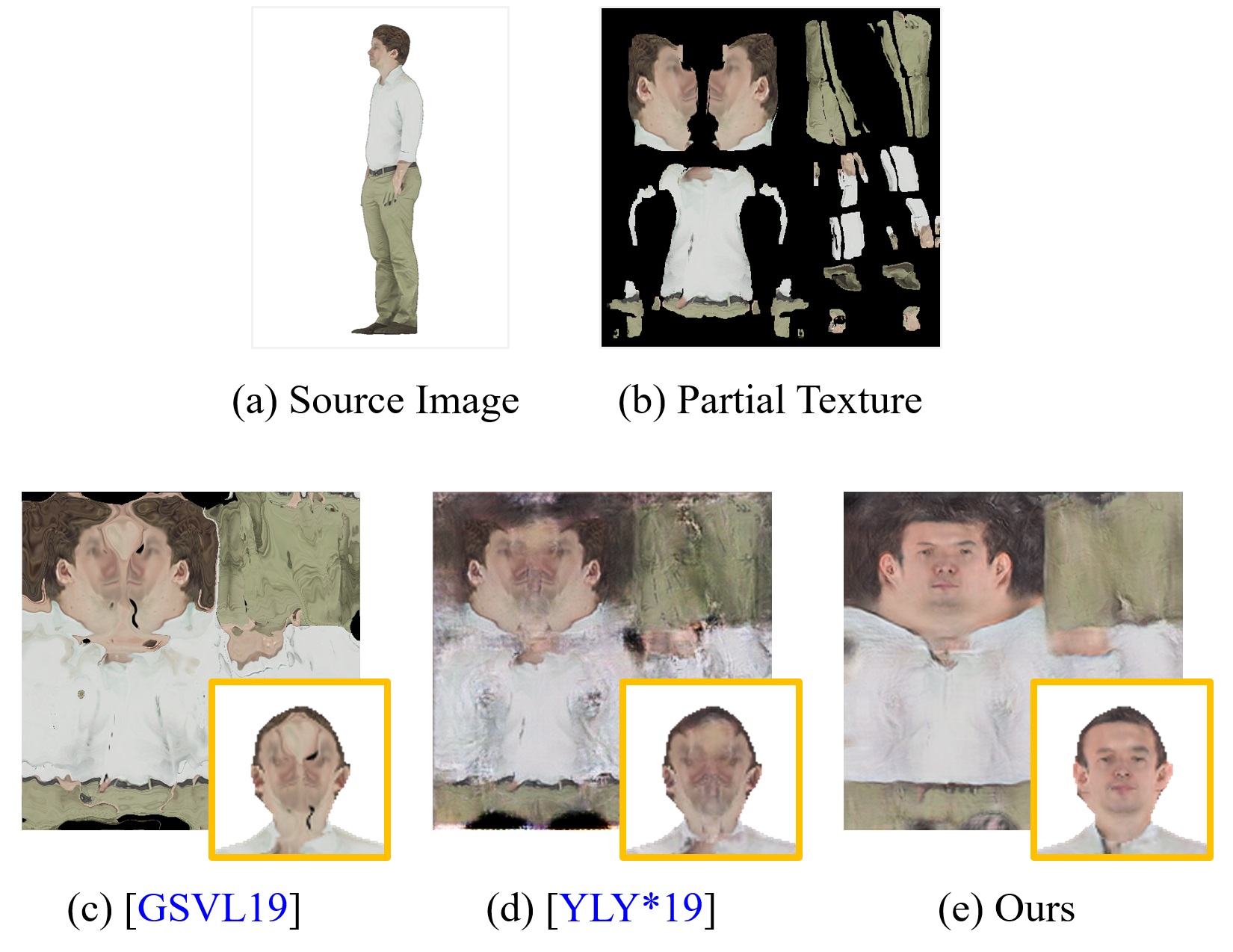}
  \caption{Generated textures using (c) coordinate-based inpainting~\cite{grigorev2019coordinate}, (d) color-based inpainting~\cite{yu2019free}, and (e) our method from (b) partial texture map. The results from inpainting methods show the preservation of the given details and structure of the partial texture map which is created from (a) source image. However, the results failed to align with the surface of the target mesh, which leads to artifacts in rendered images. 
  } 
  \label{fig:inpainting}
\end{figure}

Due to these challenges, generating a 3D human texture map cannot be simply posed as an image inpainting task~\cite{liu2018image}. Unlike image painting, where inputs and outputs are spatially aligned, the alignment of inputs and outputs is not guaranteed for 3D human texture generation tasks. Because UV alignment is essential, a process of correcting the spatial structure of the input during the 3D human texture generation task is required. The difference between the image inpainting and 3D human texture generation is highlighted in Figure~\ref{fig:inpainting}. Therefore, direct application of image inpainting methods to 3D human texture generation tends to result in a misaligned texture map. 
On the other hand, methods that utilize image-to-image translation~\cite{isola2017image} may successfully produce a texture map that is semantically aligned with the UV space of the target mesh \cite{lazova2019360, wang2019re}. 
Unfortunately, these CNN-based models tend to learn an average texture from the training data, leading to a blurry result.

In this paper, we propose a method that generates a complete human texture map from a single image while synthesizing the occluded texture with relevance to the given visible appearance. Using a neural network based on sampling and refinement strategies, our method preserves the details given in the source image in the generated texture while retaining the structural alignment with the surface of the target mesh. Similar to previous methods~\cite{alldieck2019tex2shape,lazova2019360}, we convert the source image to a partial texture map and use it as input to our method.
We also predict a 3D human mesh based on the SMPL model~\cite{loper2015smpl} from the source image using off-the-shelf method~\cite{alldieck2019tex2shape} and utilize the surface normal information of the 3D human mesh in the sampling process.
Given the partial texture map and a normal map, $SamplerNet$ completes the missing details by sampling the visible region of the texture and re-arranging them. 

The proposed sampler network overcomes the limitation of image inpainting methods~\cite{grigorev2019coordinate,yu2019free}, such as the generation of misaligned textures, by learning to align the texture to the corresponding surface of the target 3D human mesh. 

To guide the $SamplerNet$ for effective sampling, we adopt a curriculum learning scheme. Given the sampled texture map produced by $SamplerNet$, $RefinerNet$ generates a refined texture and a blending mask. The mask is used for blending the refined texture with the sampled texture to produce the final result. The blending mask helps to preserve the appearance detail presented in the source image and therefore allows to generate the final texture map with improved quality by removing artifacts from the sampled texture. Example results of the generated texture map and rendered images are shown in Figure~\ref{fig:teaser}. We conducted a set of experiments to show that our approach outperforms the various baseline methods~\cite{isola2017image,wang2018high,lazova2019360,xu2021texformer,albahar2021pose} in reproducing the details present in the source image and aligning textures to the surface of the target 3D human mesh.

\section{Related Work}

\paragraph{3D Human Texture Generation with Multi-View Images}
There have been many studies that allow to generate 3D human avatars with texture from multi-view images \cite{alldieck2018detailed, alldieck2018video, bhatnagar2019multi, zhi2020texmesh, mir2020learning, bhatnagar2019multi}. These methods project the given multi-view images or video frames back to the predicted mesh to create the partial texture maps and combine them with blending techniques to produce the final texture map. However, with only a single view, it is not clear how to extend the methods to capture all the detailed information and fill in the invisible regions.

\paragraph{3D Human Texture Generation from a Single Image} \label{sec:single_image}
Progress in generating 3D human avatars from a single image has shown a big leap in reconstructing body shapes and poses as much as that interest. However, synthesizing a texture that maintains given appearance details and restoring the occluded region with the relevance of the visible region is still challenging.
Similar to 3D human texture generation with multi-view images, some methods~\cite{natsume2019siclope,saito2019pifu,huang2020arch,he2021arch++} acquire the texture by 
predicting the back view from the given frontal view image followed by projecting the both front and back view images back to the predicted 3D human mesh. While these methods are effective in utilizing the visible textures in the source image, the results are often blurry and poorly reconstructed in the occluded area.

Some studies~\cite{grigorev2019coordinate,lazova2019360} employ a predefined mapping process that converts the visible human appearance in the source image to the UV space of the SMPL model using DensePose~\cite{guler2018densepose}. Lazova et al.~\cite{lazova2019360} utilized the partial texture map produced by a predefined mapping and takes an image-to-image translation approach to generate a full texture map. Because this method learns to generate the missing information based on training data, the generated results can often be blurry and detailed texture patterns are not recovered well when the given garment texture is unseen from the training data.

Instead of using a texture explicitly for supervision, some methods learn the human texture generation in an unsupervised manner, directly producing a texture map from the given image~\cite{wang2019re,xu2021texformer,chang2022texture}. 
These methods can generate a texture map directly from the source image by eliminating the pose variant features and minimizing the identity difference. Xu et al.~\cite{xu2021texformer} maps the given image to the texture space by predicting flow field and blends it with the generated texture map to eliminate the artifacts and maintain detailed appearance. Inspired from this approach, our method also incorporates blending process with sampled texture and refined texture.

\paragraph{Pose-guided Image and Video Synthesis}
The goal of pose-guided image synthesis is to transfer the person's appearance from a source image to the desired pose. 
Some approaches~\cite{sarkar2020neural,sarkar2021humangan,sarkar2021style} utilize the UV space of the SMPL model as an intermediate representation to achieve the task. These methods generate the latent features, which lie in the SMPL's UV space, and use them to synthesize the image with the given target pose. Other methods~\cite{neverova2018dense,liu2020neural,yoon2021pose} directly predict the texture in the UV space to transfer the pose of the source image to the target pose. However, the predicted texture map is used as a reference and undergoes a post-processing step in the image space, which does not consider the alignment with the 3D human mesh. Therefore, the predicted texture may not align with the surface of the target 3D mesh.

Instead of using color pixels from the source image, some methods~\cite{grigorev2019coordinate,albahar2021pose} map the pixel coordinates of the visible region in the image to the UV space of the SMPL model. The mapped coordinates are then inpainted and used for sampling the source image to create a full texture map. These methods reported better outcomes in retaining the local details presented in the source image compared to the methods~\cite{neverova2018dense} that directly utilize the color pixels of the source image. 
Similar to these approaches~\cite{grigorev2019coordinate,albahar2021pose}, we use a sampler network to sample the visible regions in the source image to create the texture map. Because the texture map is used for intermediate representation, it is prone to artifacts such as stretching-out or distortion when directly used for rendering as shown in Figure~\ref{fig:vis_extreme_case}.
To overcome this and improve quality, we use a texture refinement network that adjusts and refines the produced texture. Detailed illustration and evaluation with previous methods are presented in Section~\ref{experiment}.

\paragraph{Curriculum Learning} 
After the introduction of curriculum learning by Bengio et al.~\cite{bengio2009curriculum}, the strategy has been adapted to various tasks, such as language modeling~\cite{graves2017automated}, object detection~\cite{wang2018weakly}, and person re-identification~\cite{ma2017self}. We adapt the strategy of curriculum learning by progressively increasing the level of difficulty of the texture sampling task. We train $SamplerNet$ from the simplest case of the mapping process and gradually apply geometric augmentation. Through this process, the model learns to complete the texture that is aligned with the surface of the target 3D human mesh while reproducing the detailed appearance presented in the source image.


\begin{figure*}[t]
\centering
  \includegraphics[width=\linewidth]{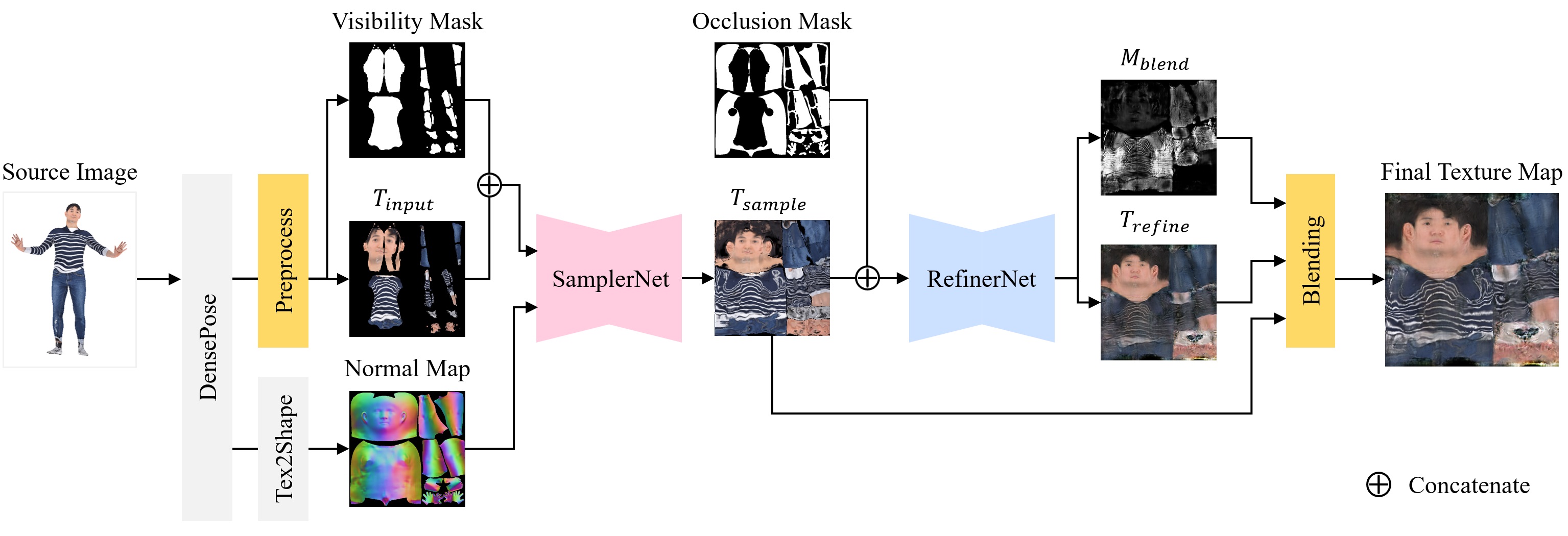}
  \caption{
    Overview of the proposed method. The source image is processed to create partial texture, visibility mask, and normal map which are given as an input to $SamplerNet$. $SamplerNet$ predicts a sampling grid that is used for producing the sampled texture. $RefinerNet$ receives the sampled texture and occlusion mask as input, and generates a refined texture and blending mask. The final output is produced by alpha blending the sampled texture with the refined texture using the blending mask.
  }
  \label{fig:model_overview}
\end{figure*}

\section{Methods}
In this section, we describe the proposed networks and training process. 
We first preprocess the human appearance in the source image to be mapped into the UV space of a 3D human mesh~\cite{loper2015smpl} and produce a partial texture map and a corresponding visibility mask. 
From the partial texture map, we obtain the geometry information using Tex2shape~\cite{alldieck2019tex2shape}, which generates the 3D human mesh from a single image by predicting the surface normal and vertex displacements in the UV space of the SMPL model.
Given the partial texture map, visibility mask, and normal map, $SamplerNet$ samples the missing appearance details from the visible regions of the source image and completes the texture map. In the following process, $RefinerNet$ generates a blending mask and a refined texture. Insufficient details and distortion artifacts in the sampled texture produced by $SamplerNet$ are adjusted in the refined texture. The final texture is then generated by blending the sampled texture with the refined texture using the blending mask. An overview of the proposed method is shown in Figure~\ref{fig:model_overview}.

\subsection{Preprocessing}\label{sec:preprocess}

Similar to previous methods~\cite{lazova2019360, alldieck2019tex2shape, grigorev2019coordinate, albahar2021pose}, we map the source image into the UV space of the SMPL model. We denote the mapped partial texture as $T_{source}$. 
Following Albahar et al.~\cite{albahar2021pose}, we combine $T_{source}$ and $T_{source}^{mirror}$, a semantically mirrored texture map of $T_{source}$, to produce a symmetric texture map $T_{input}$. $T_{input}$ is expressed as follows:
\begin{equation}\label{partial}
    T_{input} = T_{source} + (T_{source}^{mirror} \odot (1-M_{source})), 
\end{equation}
where $\odot$ is the Hadamard product. $M_{source}$ is a binary mask in which $1$ indicates a valid pixel in $T_{source}$ and $0$ indicates an invalid pixel.

For the mapping, we use the indexed UV coordinates (IUV) predicted by DensePose~\cite{guler2018densepose} to map the source image to the UV space of the SMPL model. The IUV establishes the correspondence between the human appearance in the image and the surface of the SMPL model. The surface is labeled with an index which indicates a predefined body part. Using the IUV and a lookup table provided by DensePose, the visible human appearance in the source image can map to the UV space. 
Because DensePose divides the surface of the SMPL model to exploit the left-right symmetry, the mirrored texture map $T_{source}^{mirror}$ can be created by switching the index for the IUV of the symmetrical body parts during the mapping process.

\subsection{Sampler Network}\label{sec:samplernet}
Given $T_{input}$, visibility mask, and normal map as input, $SamplerNet$ predicts a sampling grid to produce a complete texture map, $T_{sample}$. The visibility mask is a binary mask that indicates the valid pixel in $T_{input}$. Using the predicted sampling grid, the visible region of $T_{input}$ is sampled to synthesize the occluded region in $T_{input}$. Furthermore, the texture in the visible region is re-sampled to be structurally aligned to match the semantic meaning of the UV space. $SamplerNet$ consists of two encoders and a single decoder which is similar to the network proposed in Yoon et al.~\cite{yoon2021pose}. The geometry and appearance features are extracted from each dedicated encoder and are fed to the decoder. The appearance features extracted from $T_{input}$ and visibility mask at each layer of the encoder are skip-connected to the corresponding layer in the decoder. 

Both encoders used for $SamplerNet$ consist of one convolutional layer followed by five layers of residual blocks. The decoder uses the same number of residual block layers with upsampling and one convolutional layer at the end. All activations are LeakyReLU with instance normalization. The residual blocks consist of two convolutional layer with gated convolutions~\cite{yu2019free}.

When training $SamplerNet$, we observed that $SamplerNet$ often fails to preserve the given appearance detail in $T_{input}$. To overcome this, we apply the curriculum learning strategy. Before describing the details of the proposed curriculum learning scheme, we will first address the data preparation process for the training.

\subsubsection{Training Data Preparation}\label{sec:Training_Data}
$SamplerNet$ is trained with a curriculum learning scheme, which begins from the simplest case where the given input texture is perfectly aligned with the UV space, to the hard case where the input texture is misaligned with the UV space. For the simplest case, if the predicted IUV perfectly aligns with the surface coordinates of the SMPL model, partial texture $T_{source}$ can be acquired by masking out the ground truth texture map $T_{GT}$ with $M_{source}$:
\begin{equation}\label{ideal}
    T_{source} = T_{GT} \odot M_{source}.
\end{equation}
We now denote $T_{source}$ obtained by Equation~\ref{ideal} as $T_{GT}^{M}$.

Due to imperfect prediction, however, DensePose often fails to locate the exact pixel position that corresponds to the geometric position. This prevents the partial texture from being semantically aligned with $T_{GT}$. 
We approximate this misalignment using a geometric transformation function $f(\cdot)$, and thus $T_{source}$ can be expressed as augmentation as follows:
\begin{equation}\label{approx}
\begin{split}
    T_{source} \approx T_{Augment} = f(T_{GT}^{M}, \alpha),
\end{split}
\end{equation}
where $\alpha$ is a control parameter for $f(\cdot)$. 
By changing $\alpha$, interpolation from $T_{GT}^{M}$ to $T_{source}$ is approximated which enables the curriculum learning scheme.

\subsubsection{Region-wise Augmentation} 
\begin{figure}[t]
\centering
  \includegraphics[width=\linewidth]{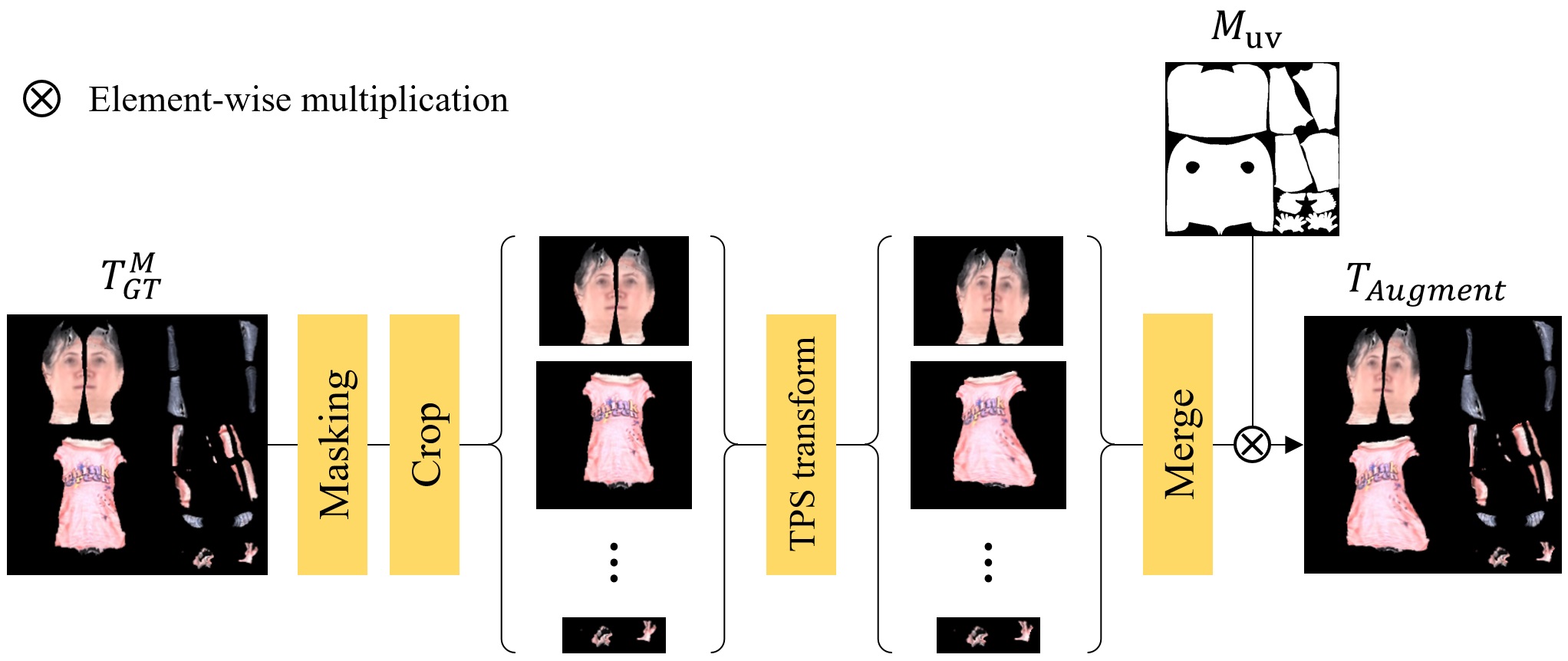}
  \caption{
      Illustration of the region-wise augmentation.
  } \label{fig:region-wise_augmentation}
\end{figure}

For the augmentation $f(\cdot)$, we use the thin-plate-spline (TPS) transformation in a region-wise manner (Figure~\ref{fig:region-wise_augmentation}). There are two reasons why naive augmentation techniques are inadequate for texture generation. First, unlike general images, texture maps have a unique structure that corresponds to a 3D geometry. Second, although $T_{source}$ is non-linearly deformed from $T_{GT}$, the deformation is limited to the UV structure of the SMPL model. Therefore, augmentation should be performed in consideration of this structure.

We divide the UV space into six different regions which correspond to the head, body, legs, arms, feet, and hands. 
We crop each region with the bounding box and apply the TPS transform individually.
To prevent unintentional cropping, each body part is masked before the process.
Then, we merge all of the transformed regions back to form a single texture followed by multiplying it by the UV mask, $M_\text{uv}$, to produce the final deformed texture for training. $M_\text{uv}$ is a binary mask, which represents valid UV coordinates in the UV space. 
For the transformation, we assign control points to each region and shift these points with a random value determined from a uniform distribution $\textit{U}(0, \alpha)$. $\alpha$ is expressed as follows:
\begin{equation}\label{augment_amount}
    \alpha = \begin{cases}
                0,            & step=0, \\
                0.1 + (step * \delta), & step>0. \\
                \end{cases}
\end{equation}
$step$ indicates the current curriculum step and $\delta$ is a hyper-parameter.

\subsubsection{Curriculum Learning}\label{method:cl}
The goal of curriculum learning is to guide $SamplerNet$ to maintain the given appearance detail in $T_{input}$ while enforcing the semantic alignment. To this end, we divide the training into two curriculum steps: sampling the occluded texture from the visible region and re-arranging the given visible region. First we will describe the data preparation for the curriculum learning.

In the initial step, $step=0$, the objective is to complete the given partial texture map without considering the alignment in the visible region. Hence, $\alpha$ in Equation~\ref{approx} is set to $0$, which makes $f(\cdot)$ an identity mapping function, resulting in $T_{Augment}$ identical to $T_{GT}^M$. As $T_{GT}^M$ is semantically aligned with the UV space, $SamplerNet$ is encouraged to sample for the missing regions only. A single $step$ is set to $4,000$ iterations, which is sufficient for the model to see the whole training data set twice.

After the initial step, $\alpha$ is set to $0.1$ and increased by $\delta$ after every single step. Here, the goal is to encourage the model to begin re-sampling the visible region to enforce the structural alignment.

For the steps equal to or greater than 3, we additionally use the partial texture produced by DensePose to reduce the domain gap between training data and inference data.

The data used in the curriculum learning can be expressed as follows: 
\begin{align}
T_{source} = \begin{cases}
            T_{Augment},                               & step < 3, \\
            T_{Augment} \; \text{or} \; T_{DensePose}, & step \geq 3. \\
            \end{cases}
\label{curriculum}
\end{align}
where $T_{DensePose}$ is a partial texture produced by DensePose.
The variation of the partial texture is visualized in the supplementary material.

The effect of using this augmentation is verified in Section~\ref{ablation}. With the curriculum learning and the region-wise augmentation, $SamplerNet$ effectively samples a partial texture map to synthesize all the occluded region while retaining the given visible part and aligning texture to the surface of the target 3D mesh.

\subsubsection{Loss Functions}

To enforce the alignment and proper sampling, we train $SamplerNet$ by minimizing a reconstruction loss and a perceptual loss between $T_{sample}$ and $T_{GT}$.

The reconstruction loss is expressed as follows:
\begin{equation}
    \mathcal{L}_{Recon} = \sum_{i=1}^N || w_i \cdot M_\text{body}^i \odot (T_{sample} - T_{GT}) ||_1,
\end{equation}
where $M_\text{body}^i$ is a set of binary masks for the six body parts in the UV space and $w_i$ is its corresponding weights. $w_i$ is set to ${6,1,1,1,1,\text{and }1}$ for each face, body, leg, arm, foot, and hand.

For the perceptual distance, we use LPIPS~\cite{zhang2018unreasonable}, which can be expressed as follows:
\begin{equation}
    \mathcal{L}_{LPIPS} = LPIPS(T_{sample}, T_{GT}),
\end{equation}
\noindent
where $LPIPS(\cdot)$ extracts features from the two inputs using AlexNet~\cite{krizhevsky2014one} and calculates the cosine distance between the extracted features.

The total loss function for $SamplerNet$ is expressed as follows:
\begin{equation}
    \mathcal{L}_{Sampler} = \lambda_{Recon}\mathcal{L}_{Recon} + \lambda_{LPIPS}\mathcal{L}_{LPIPS},
\end{equation}
\noindent
where $\lambda_{Recon}$ and $\lambda_{LPIPS}$ are set to $1$ and $1$, respectively.

\subsection{Refiner Network}\label{sec:refinenet}
\begin{figure}[t]
\centering
\includegraphics[width=\linewidth]{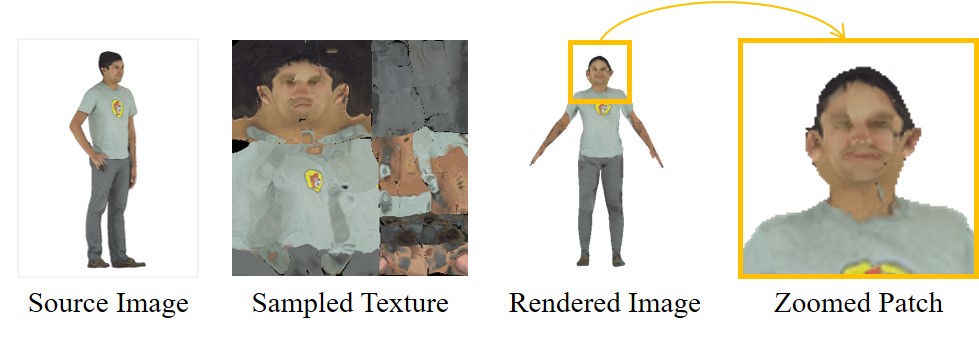}
\caption{Artifact due to the $SamplerNet$'s prediction error}
\label{fig:vis_extreme_case}
\end{figure}

After $SamplerNet$ completes the partial texture, the resulting texture map is refined. 
$SamplerNet$ implicitly learns to sample the missing texture information from the visible region by following the guidance of given geometry information and minimizing the loss in the training process. 
During the process of sampling the given information according to the geometry information, some artifacts are accompanied as shown in Figure~\ref{fig:vis_extreme_case}.

$RefinerNet$ alleviates this problem by refining the details in $T_{sample}$. $RefinerNet$ employs a U-Net-like architecture with three down and up sampling layers, and 9 residual blocks for the bottleneck. $RefinerNet$ receives $T_{sample}$ and occlusion mask as input and produces a refined texture map $T_{refine}$ and a blending mask $M_{blend}$. The occlusion mask is a binary mask, which is acquired by subtracting $M_{src}$ from $M_\text{uv}$.

As described in Section~\ref{sec:single_image}, Xu et al.~\cite{xu2021texformer} proposed a mask-fusion strategy that blends the two texture outputs using a predicted mask to preserve the fine texture details. 
Inspired by this, we adapt a texture blending process that exploits details in $T_{sample}$. 
The blending mask helps to preserve the given appearance detail by replacing the artifacts in the sampled texture with the refined texture. The final output is computed as follows:
\begin{equation}
  T_{final} = T_{sample} \odot M_{blend} + T_{refine} \odot (1- M_{blend}).
\end{equation}

Despite the well reconstructed human appearance features, the fine details observed in $T_{sample}$ can sometimes be lost and become blurry in $T_{refine}$. The blending of $T_{refine}$ and $T_{sample}$ using $M_{blend}$ leverages the advantages of each generated texture. 
The effect of texture blending is demonstrated in Figure~\ref{fig:blending_texture}.

\subsubsection{Loss Functions} 
To produce a texture map of perceptually plausible quality with minimal artifacts, we minimize the following objective terms when training $RefinerNet$:
\begin{equation}
\begin{split}
  \mathcal{L}_{Refiner} = \lambda_{Recon}\mathcal{L}_{Recon} + \lambda_{VGG}\mathcal{L}_{VGG} \\
  + \lambda_{GAN}\mathcal{L}_{GAN} + \lambda_{FM}\mathcal{L}_{FM},
\end{split}
\end{equation}
where $\mathcal{L}_{Recon}$, $\mathcal{L}_{VGG}$, $\mathcal{L}_{GAN}$, and $\mathcal{L}_{FM}$ are the reconstruction loss, perceptual loss, adversarial loss, and feature matching loss, respectively. 

\begin{figure}[t]
\centering
  \includegraphics[width=\linewidth]{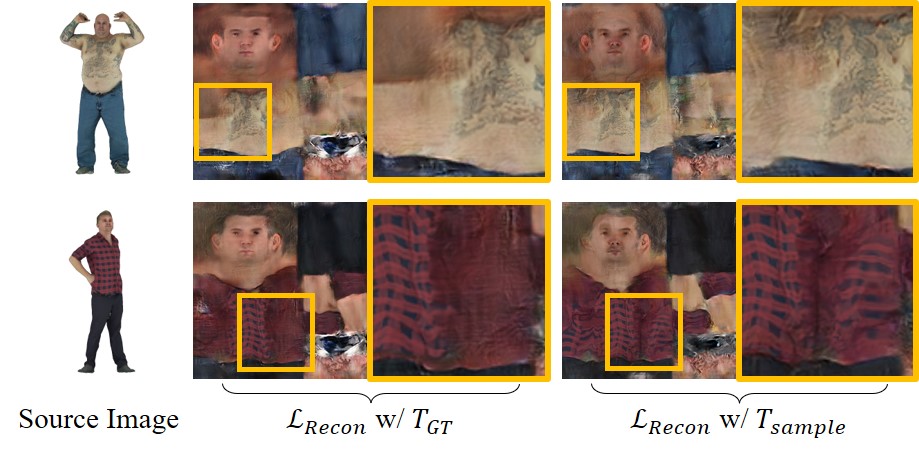}
  \caption{Visual comparison of applying the reconstruction loss to $T_{sample}$ and the ground truth texture map. 
  }
  \label{fig:ablation_reconstruction}
\end{figure}

For $\mathcal{L}_{Recon}$, instead of calculating the loss between $T_{final}$ and $T_{GT}$, we calculate the loss between $T_{final}$ and $T_{sample}$. Because there can be multiple texture maps corresponding to one source image, calculating the loss directly from the ground truth map is restrictive. For example, in the source image, a person wearing a T-shirt with a pattern on the front, may have the same pattern, different pattern, or even no pattern on the back. Thus, a pixel-wise loss with $T_{GT}$ will lead to a texture with the averaged color output (Figure~\ref{fig:ablation_reconstruction}). We used the loss between $T_{final}$ and $T_{sample}$ to guide the model in a direction that more respects the estimated $T_{sample}$.
$\mathcal{L}_{Recon}$ is expressed as follows:
\begin{equation} \label{recon_loss}
  \mathcal{L}_{Recon} = || T_{sample} - T_{final} ||_1.
\end{equation}

For $\mathcal{L}_{VGG}$, we use the pre-trained VGG-19~\cite{simonyan2014very} to calculate the perceptual distance between $T_{final}$ and $T_{GT}$ by extracting features from each layer $l$ as performed in Wang et al.~\cite{wang2018high}. $\mathcal{L}_{VGG}$ is expressed as follows:
\begin{equation} \label{percept_loss}
\begin{split}
 \mathcal{L}_{VGG} = \sum_{i=1}^N {\frac{1}{w_i}} ||VGG_{i}(T_{GT}) - VGG_{i}(T_{final}) ||_1.
\end{split}
\end{equation}
Here, $VGG_i$ denotes the layer of the VGG-19 network, where $i \in \{1,6,11,20,29\}$. $w_i$ is set to $32,16,8,4, \text{and } 1$ for each layer.

We use an adversarial loss with the PatchGAN discriminator~\cite{isola2017image}. The objective function can be expressed as follows:
\begin{equation} \label{GAN_loss}
\begin{split}
 \mathcal{L}_{GAN} = \mathbb{E}_{T_{GT}}[\log{D(T_{GT})}]  + \mathbb{E}_{T_{sample}}[\log{(1 - D(T_{final})}],
\end{split}
\end{equation}
where $D$ denotes the discriminator. We additionally use a feature matching loss $\mathcal{L}_{FM}$~\cite{wang2018high}, which minimizes the feature distance between $T_{final}$ and $T_{GT}$ of the discriminator to stabilize the training. $\mathcal{L}_{FM}$ is expressed as follows:
\begin{equation} \label{fm_loss}
\begin{split}
 \mathcal{L}_{FM} = \sum_{i=1}^N ||D_{l_i}(T_{GT}) - D_{l_i}(T_{final}) ||_1,
\end{split}
\end{equation}
where $l_i$ represents a set of layers after the activation function and $i \in \{1,2,3\}$. The weights used in the total loss function are $\lambda_{Recon}=10$, $\lambda_{VGG}=10$, $\lambda_{GAN}=1$, and $\lambda_{FM}=10$.

\subsection{Training Details} 
We used the Adam optimizer~\cite{kingma2014adam} with a learning rate of $0.0002$ and beta parameters set to $0.9$ and $0.999$ for both networks, $SamplerNet$ and $RefinerNet$. The region-wise augmentation for training $SamplerNet$ is applied with the probability of $0.8$ and $\delta$ is set to $0.025$. 
We used color augmentation for training $RefinerNet$ with the probability of $0.5$. The batch size is set to 8, and each model is trained separately for 30,000 iterations on an NVIDIA GTX 1080 Ti GPU.


\begin{figure*}[!htp]
\centering
  \includegraphics[width=\textwidth]{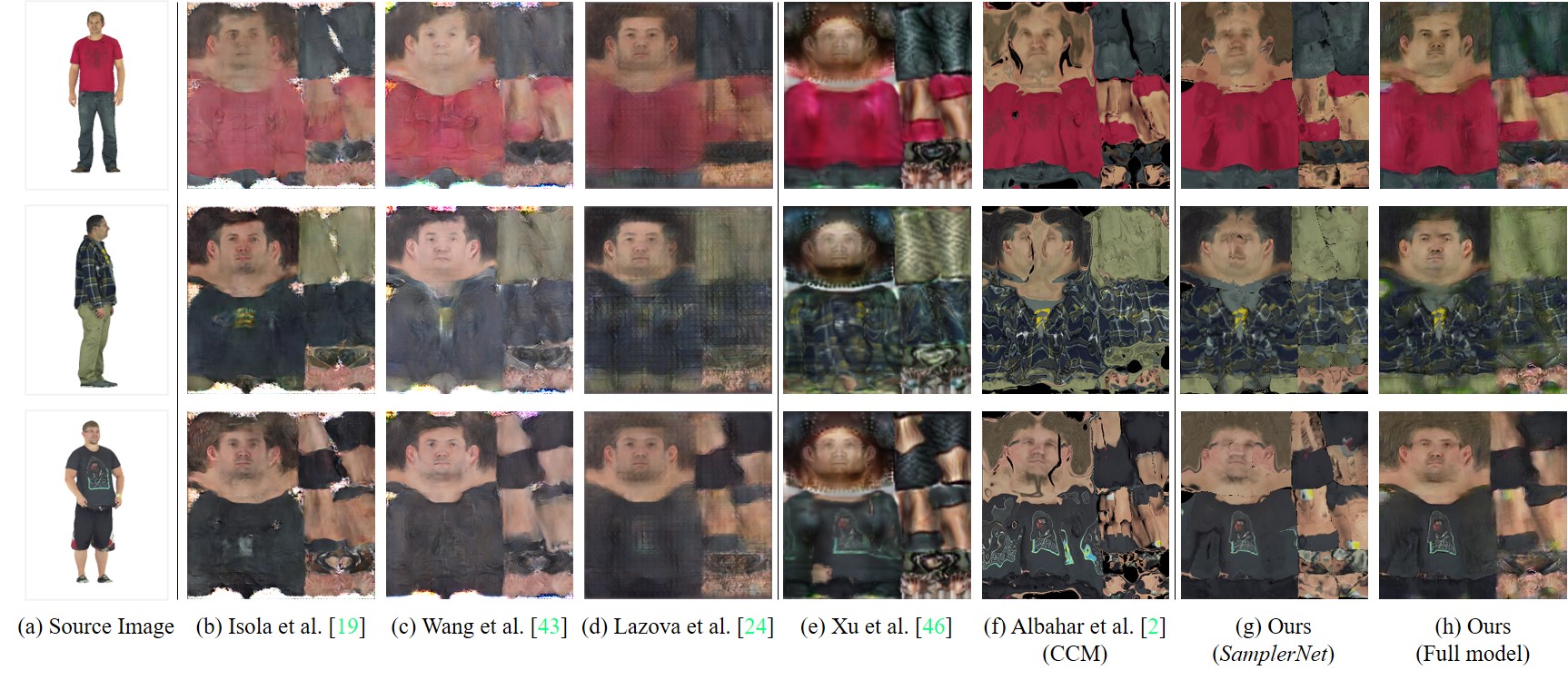}
  \caption{Visual comparison with previous methods using Digital Wardrobe~\cite{bhatnagar2019multi} dataset. The texture maps are generated for the person viewed from randomly rotated perspectives in the horizontal direction.}
\label{fig:qualitative_texture}
\end{figure*}

\begin{figure*}[!ht]
\centering
  \includegraphics[width=\textwidth]{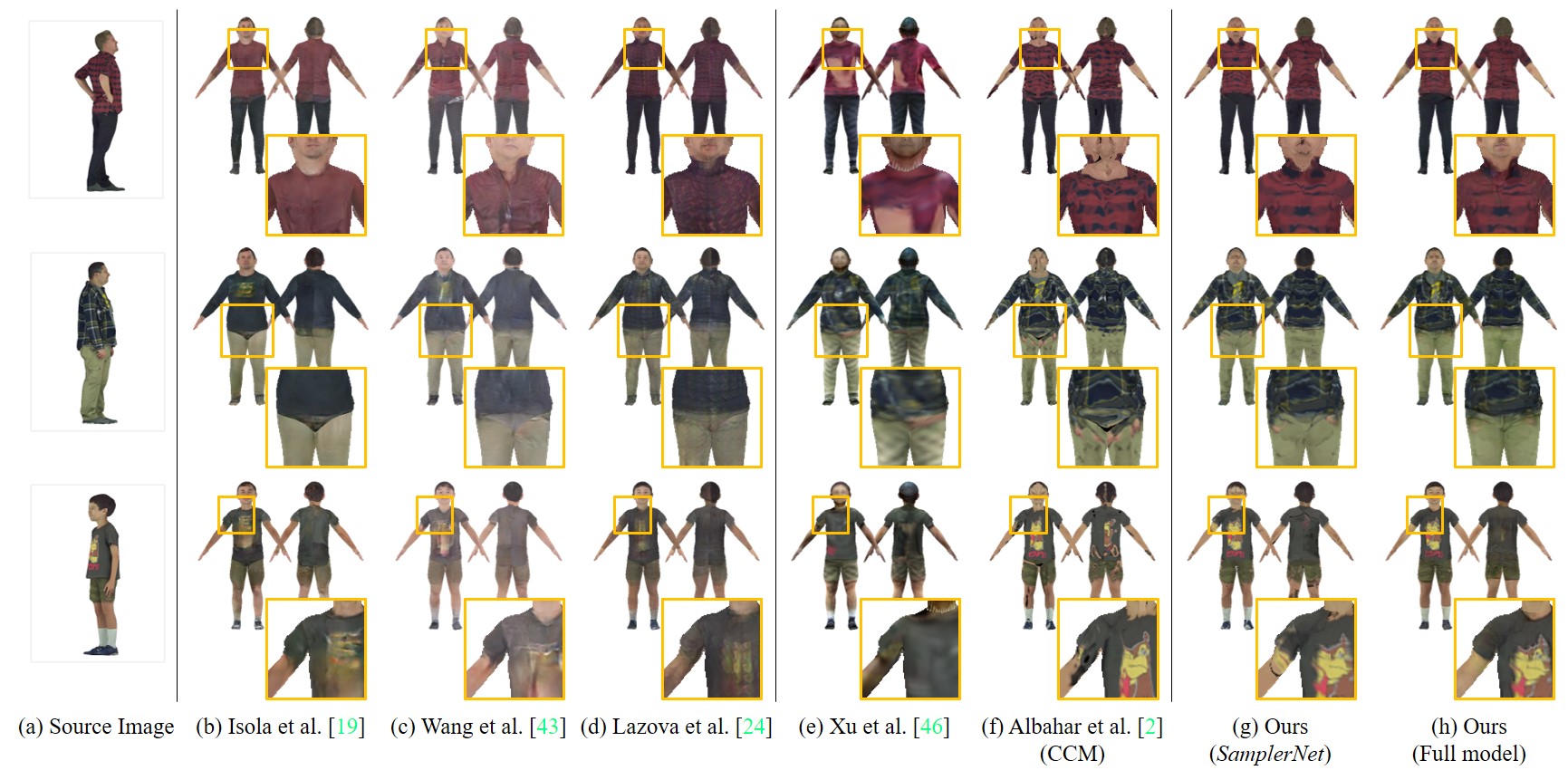}
  \caption{Visual comparison with previous methods using Digital Wardrobe~\cite{bhatnagar2019multi} dataset. The images are rendered with texture maps that are generated using the person viewed from the randomly rotated perspectives in the horizontal direction. We used Tex2Shape~\cite{alldieck2019tex2shape} to estimate the target mesh from the source image.}
\label{fig:qualitative_rendered}
\end{figure*}

\begin{table*}[!ht] 
\begin{center}
\caption{Quantitative evaluation of the generated texture map. The texture maps were generated using the input images with various views. For the *Avg, we used all the images in the range of [$-90^{\circ},90^{\circ}$] with $10^{\circ}$ interval as input and averaged the calculated scores from the generated texture maps. }
\resizebox{\textwidth}{!}{%
    \begin{tabular}{l|ccc|ccc|ccc}
    \noalign{\smallskip}
    \hline
    Angle 
    &\multicolumn{3}{c|}{$0^{\circ}$ (front view)} 
    &\multicolumn{3}{c|}{$90^{\circ}$ (side view)}   
    &\multicolumn{3}{c}{*Avg} \\
    \hline 
    Method 
    & LPIPS↓ & PSNR↑ & SSIM↑  
    & LPIPS↓ & PSNR↑ & SSIM↑  
    & LPIPS↓ & PSNR↑ & SSIM↑   \\
    \hline
    Isola et al.~\cite{isola2017image}
    & 0.3142 & 18.49 & 0.5305 
    & 0.3163 & 18.48 & 0.5309 
    & 0.3127 & 18.60 & 0.5314 \\
    Wang et al.~\cite{wang2018high}
    & 0.3085 & 18.21 & 0.5323 
    & 0.3115 & 18.40 & 0.5356 
    & 0.3078 & 18.34 & 0.5334 \\
    Lazova et al.~\cite{lazova2019360}
    & 0.3047 & \textbf{18.91} & 0.5604 
    & 0.3180 & \textbf{18.73} & 0.5522 
    & 0.3075 & \textbf{18.97} & 0.5561 \\
    \hline
    Xu et al.~\cite{xu2021texformer}
    & 0.3713 & 15.70 & 0.4941 
    & 0.3777 & 15.35 & 0.4948 
    & 0.3755 & 15.54 & 0.4929 \\
    Albahar et al.~\cite{albahar2021pose} (CCM)
    & 0.2884 & 16.57 & 0.5144
    & 0.2700 & 17.40 & 0.5336 
    & 0.2804 & 17.15 & 0.5182 \\
    \hline
    Ours ($SamplerNet$)
    & 0.2435 & 17.53 & 0.5501 
    & 0.2445 & 17.84 & 0.5620 
    & 0.2448 & 17.78 & 0.5509 \\ 
    Ours ($SamplerNet+RefinerNet$)
    & \textbf{0.2230} & 18.04 & \textbf{0.5853} 
    & \textbf{0.2227} & 18.62 & \textbf{0.5988} 
    & \textbf{0.2236} & 18.26 & \textbf{0.5837} \\
    \hline
    \noalign{\smallskip}
    \end{tabular} \label{table:quantitative_texturespace}
}
\end{center}
\end{table*}

\begin{table*}[!ht] 
\begin{center}
\caption{Quantitative evaluation of the rendered image. We compared the methods using the input images with various views and rendered the generated texture maps on the target 3D mesh. We used Tex2Shape~\cite{alldieck2019tex2shape} to estimate the target mesh from the input image and applied it for rendering in all methods. For the *Avg, we used all the rendered images in the range of [$-90^{\circ},90^{\circ}$] with $10^{\circ}$ interval as input and averaged the calculated scores.
}
\resizebox{\textwidth}{!}{%
    \begin{tabular}{l|ccc|ccc|ccc}
    \noalign{\smallskip}
    \hline
    Angle 
    &\multicolumn{3}{c|}{$0^{\circ}$ (front view)} 
    &\multicolumn{3}{c|}{$90^{\circ}$ (side view)}   
    &\multicolumn{3}{c}{*Avg} \\
    \hline
    Method 
    & CosSim-A↑ & CosSim-I↑ & LPIPS↓ 
    & CosSim-A↑ & CosSim-I↑ & LPIPS↓ 
    & CosSim-A↑ & CosSim-I↑ & LPIPS↓ \\
    \hline
    Isola et al.~\cite{isola2017image}
    & 0.7421 & 0.7685 & 0.2017
    & 0.7088 & 0.7405 & 0.2026 
    & 0.7294 & 0.7577 & 0.1999 \\
    Wang et al.~\cite{wang2018high}
    & 0.7266 & 0.7542 & 0.2058
    & 0.6978 & 0.7279 & 0.2063 
    & 0.7155 & 0.7430 & 0.2035 \\
    Lazova et al.~\cite{lazova2019360}
    & 0.7510 & 0.7786 & 0.1965
    & 0.7199 & 0.7452 & 0.2024
    & 0.7384 & 0.7642 & 0.1967 \\
    \hline
    Xu et al.~\cite{xu2021texformer}
    & 0.7480 & 0.7711 & 0.1923
    & 0.7338 & 0.7401 & \textbf{0.1980}
    & 0.7421 & 0.7578 & 0.1952  \\
    Albahar et al.~\cite{albahar2021pose} (CCM)
    & 0.7638 & 0.7881 & 0.1944
    & 0.7468 & 0.7532 & 0.2007
    & 0.7592 & 0.7751 & 0.1952 \\
    \hline
    Ours ($SamplerNet$)
    & 0.7673 & \textbf{0.7944} & 0.1931
    & 0.7500 & \textbf{0.7555} & 0.2007
    & 0.7609 & \textbf{0.7779} & 0.1954 \\ 
    Ours ($SamplerNet+RefinerNet$)
    & \textbf{0.7678} & 0.7938 & \textbf{0.1868}
    & \textbf{0.7504} & 0.7532 & 0.1990 
    & \textbf{0.7615} & 0.7759 & \textbf{0.1912}  \\ 
    \hline
    \noalign{\smallskip}
    \end{tabular} \label{table:quantitative_imapespace}
}
\end{center}
\end{table*}

\section{Experiments}\label{experiment}

\begin{figure}[ht]
\centering
  \includegraphics[width=\linewidth]{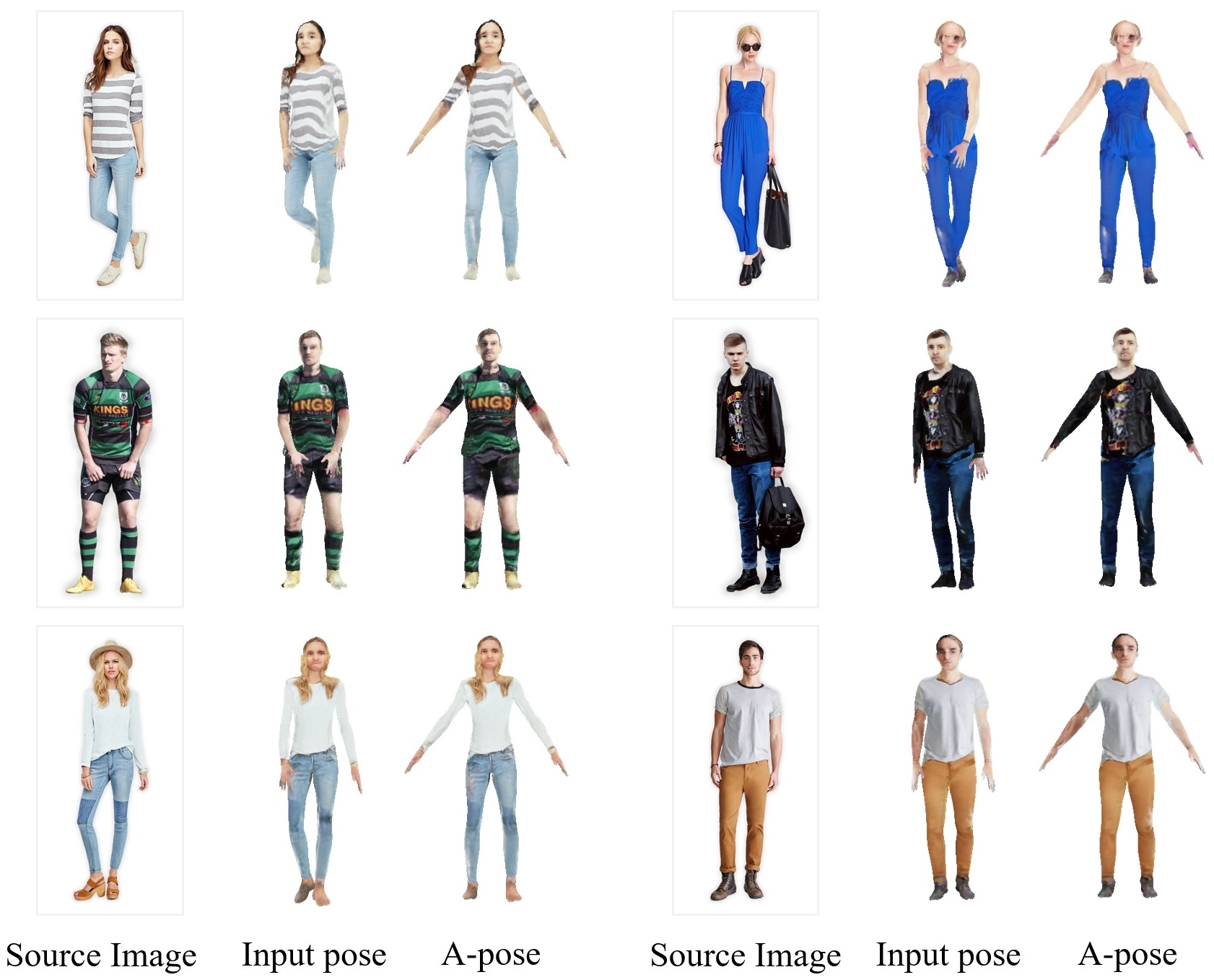}
  \caption{
      Rendered images of 3D human avatars produced using real human images. The images are from SHHQ~\cite{fu2022styleganhuman} dataset. The 3D human pose and the camera parameters were predicted from the images using RSC-Net~\cite{xu20213d}.
  }
\label{fig:SHHQ}
\end{figure}

In this section, we compare our method with previous approaches. Furthermore, several experiments were carried out to evaluate the effect of region-wise augmentation, curriculum learning, and texture blending. A total of 1,441 texture maps were used for training the networks: $929$ from SURREAL~\cite{varol2017learning} dataset, $512$ from our newly gathered dataset. The textures were rendered with SMPL~\cite{loper2015smpl} in various human poses to generate partial texture maps for training. To obtain various poses, we randomly sampled $100$ poses in the collected animations from Mixamo~\cite{adobe2020mixamo}. The textures and poses were randomly paired and rendered in the range of $[-90^\circ, 90^\circ]$ with an interval of $10^\circ$. The texture maps were resized to $256\times256$ for both training and testing. For the evaluation, we used Digital Wardrobe~\cite{bhatnagar2019multi} dataset, which contains $96$ textures that are different from the training set.
Additionally, we generated texture using samples from SHHQ~\cite{fu2022styleganhuman} dataset to assess the generalizability of our method to real human images. The results are shown in Figure~\ref{fig:SHHQ}.

\subsection{Comparisons}

We compared our method with previous methods in both generated texture map and rendered image to quantitatively measure the quality of the generated texture map.
To evaluate the quality of the generated texture map, we used the following metrics: Structural Similarity (SSIM) \cite{wang2004image}, Peak Signal-to-Noise Ratio (PSNR), and Learned Perceptual Image Patch Similarity (LPIPS) \cite{zhang2018unreasonable}. SSIM and PSNR measure the reconstruction quality and LPIPS measures the perceptual quality using VGG-16~\cite{simonyan2014very} as a backbone.
We further evaluated the results after applying the textures to the 3D human mesh. For this, we employed LPIPS, and cosine similarity (CosSim) of features extracted from PCB network~\cite{sun2018beyond}. PCB network~\cite{sun2018beyond} is used for a person re-identification task, which aims to find an identical person from the different cameras. The similarity between the source image and the rendered image using the generated texture is high when the CosSim is close to one. 
To evaluate the quality of the synthesized texture in the occluded region, CosSim was measured in two different images: one rendered with a pose identical to that of the person in the given image and the other rendered with A-pose in the frontal view. We denote the cosine similarity to the image rendered in the identical pose as CosSim-I and to the image rendered with A-pose as CosSim-A. LPIPS was calculated with the image rendered in the identical pose.
To render the image using a generated texture, we obtained a 3D human mesh from the given image using Tex2Shape~\cite{alldieck2019tex2shape}. We also used RSC-Net~\cite{xu20213d} to predict the 3D human pose and the camera parameters from the given image.

We evaluate our method with previous approaches that can generate texture maps using image-to-image translation: Isola et al.~\cite{isola2017image}, Wang et al.~\cite{wang2018high}, and Lazova et al.~\cite{lazova2019360}. Additionally, we also compared our method with approaches that utilizes pixel coordinate information: Xu et al.~\cite{xu2021texformer} and Albahar et al.~\cite{albahar2021pose}. 
For Albahar et al.~\cite{albahar2021pose}, we exploited their coordinate completion model (CCM) as this can be used for texture generation. We used the available implementations provided by the authors except for Lazova et al.~\cite{lazova2019360}, in which we tried to replicate the implementation following the descriptions in the paper. 
All methods were trained with the same data as our methods except for Xu et al.~\cite{xu2021texformer}.
Because our data was not compatible to train the method, we used the pretrained model released with their code. The input for all baseline methods was the symmetric partial texture map which was produced using Equation~\ref{partial} except for Xu et al.~\cite{xu2021texformer} which can generate texture map directly from the given image.
To evaluate the robustness of each method against various view angles, we generated texture maps using the images viewed from different perspectives by horizontally rotating the 3D human mesh in the range of $[-90^{\circ},90^{\circ}]$.

The quantitative results for texture map and rendered image are shown in Table~\ref{table:quantitative_texturespace} and Table~\ref{table:quantitative_imapespace}, respectively. 
The methods that generate a texture map using image-to-image translation~\cite{isola2017image, wang2018high, lazova2019360} achieved high scores in PSNR. Nonetheless, the produced texture lacked the detailed patterns present in the source image and failed to align with the surface of the target 3D mesh as shown in Figure~\ref{fig:qualitative_texture}(b)-(d) and Figure~\ref{fig:qualitative_rendered}(b)-(d).

The methods that utilize pixel coordinates~\cite{xu2021texformer,albahar2021pose} tend to preserve visible texture pattern in the given image better compared to image-to-image translation methods~\cite{isola2017image, wang2018high, lazova2019360} as shown in Figure~\ref{fig:qualitative_rendered}(e)-(f). 
Xu et al.~\cite{xu2021texformer} directly generate a texture map from the given image by predicting the flow field from the source image to the UV space of the SMPL model. This allows the preservation of appearance detail. However, some artifacts are apparent when the side view image is given as shown in Figure~\ref{fig:qualitative_rendered}(e).
Albahar et al.~\cite{albahar2021pose} synthesized the texture pattern that is relevant to the given details but often with severe artifacts which defect the alignment with the surface of the target 3D mesh as shown in Figure~\ref{fig:qualitative_rendered}(f).
$SamplerNet$ was able to preserve the given details and sample the occluded texture patterns while maintaining the alignment with the surface of the target 3D mesh. This is reflected by higher values of both CosSim-A and CosSim-I reported in Table~\ref{table:quantitative_imapespace}. Given the sampled texture with improved alignment, our full method produced visually better results compared to those produced by previous methods in terms of maintaining the appearance details present in the source image and synthesizing missing details exploiting the information from the visible region as shown in Figure~\ref{fig:qualitative_rendered}(h).

\subsection{Ablation study} \label{ablation}
\subsubsection{Region-wise Augmentation} 
The goal of region-wise augmentation is to approximate the transformation caused by DensePose~\cite{guler2018densepose}. 
To evaluate the similarity of the proposed augmentation with DensePose, we compared the results from $SamplerNet$ trained with several different augmentation alternatives that are applied in a region-wise manner. The curriculum learning was excluded in this experiment to examine the effect of augmentation. $\alpha$ and the probability for the augmentation is fixed to $0.25$ and $0.999$, respectively. 
We also compared $SamplerNet$ trained with and without region-wise augmentation using TPS. For the case without it, the control points were set at an equispaced $6\times6$ grid followed by shifting them according to a random value from a uniform distribution $\textit{U}(0, \alpha)$. We used same $\alpha$ value for both.
As shown in Table~\ref{table:augmentations}, $SamplerNet$ trained with TPS in a region-wise manner resulted in the scores closest to $SamplerNet$ trained with DensePose data.

\begin{table}[ht] 
\begin{center}
\caption{Comparison between various augmentation techniques. The bold number represents the best score and the underlined number represents the score closest to that produced by the model trained with DensePose.}
\resizebox{\linewidth}{!}{%
    \begin{tabular}{l|c|ccc}
    \noalign{\smallskip}
    \hline
    Augmentation  & Region-wise & LPIPS↓  &    PSNR↑  &    SSIM↑  \\
    \hline
    \hline
    Rotate        & \checkmark  & 0.2348  &	17.99 &	0.5657 \\
    Translate     & \checkmark  & 0.2468  &	18.01 &	0.5669 \\
    TPS           & \text{\sffamily x}  & 0.2323  &	18.15 &	0.5758 \\
    TPS           & \checkmark  & \underline{0.2270}  &	\underline{18.26} &	\underline{0.5817} \\
    \hline
    DensePose     & -           & \textbf{0.2142} &	\textbf{18.86} &	\textbf{0.6071} \\
    \hline
    \noalign{\smallskip}
    \end{tabular} \label{table:augmentations}
}
\end{center}
\end{table}

\subsubsection{Curriculum Learning} 
\begin{figure}[!t]
\centering
  \includegraphics[width=\linewidth]{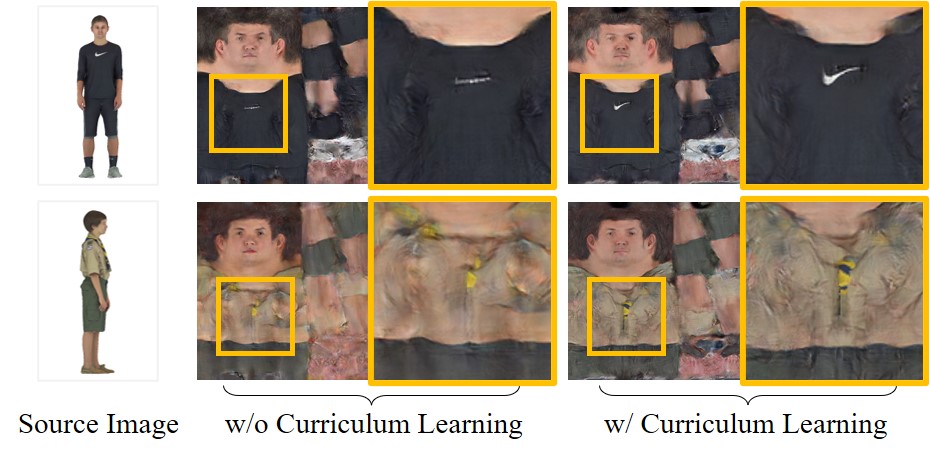}
  \caption{Visual comparison of the results produced with and without curriculum learning.
    }
\label{fig:ablation_curriculum}
\end{figure}

We evaluated the effectiveness of the curriculum learning by comparing our networks trained with and without it. The comparison results are shown in Table~\ref{table:curriculum}. The results from the networks trained with curriculum learning achieved better scores for all of the evaluation metrics. Moreover, the network trained with curriculum learning produced qualitatively superior results in terms of reproducing the given appearance as shown in Figure~\ref{fig:ablation_curriculum}.

\begin{table}[ht]
\begin{center}
\caption{Ablation study for the method trained with and without curriculum learning.}

\begin{tabular}{l|ccc}
    \noalign{\smallskip}
    \hline
    Method 
    & LPIPS↓ & PSNR↑ & SSIM↑           \\
    \hline
    \hline
    w/o curriculum learning 
    & 0.2238 & 18.09 & 0.5920          \\
    w/ curriculum learning  
    & \textbf{0.2220} & \textbf{18.18} & \textbf{0.5935} \\ 
    \hline
    \noalign{\smallskip}
\end{tabular} \label{table:curriculum}
\end{center}
\end{table}

\subsubsection{Texture Blending} 
\begin{figure}[ht]
\centering
  \includegraphics[width=\linewidth]{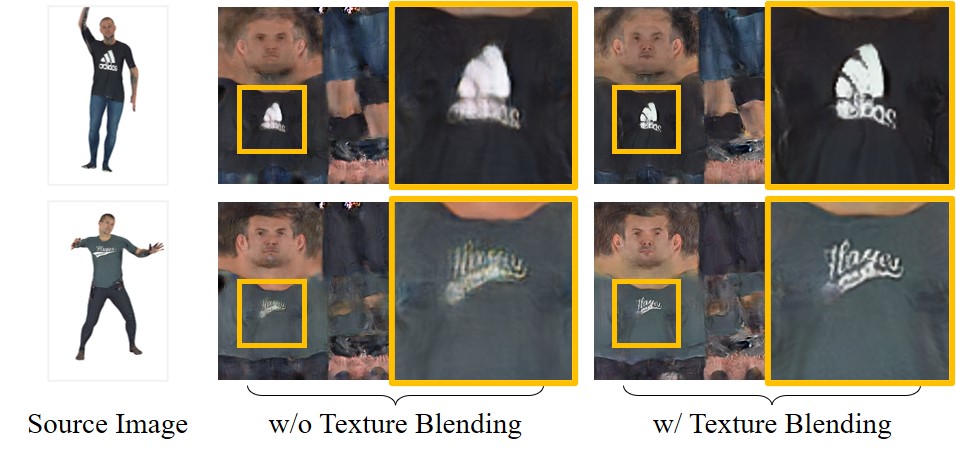}
  \caption{Visual comparison of the resulting texture maps from the model trained with and without using the blending mask.}
  \label{fig:blending_texture}
\end{figure}

We evaluate the effect of blending the refined texture with the sampled texture using a blending mask. We trained $RefinerNet$ with two different settings: producing the final texture map using the blending mask and directly generating the final texture map. 
The results are shown in Table~\ref{table:blending_texture} and Figure~\ref{fig:blending_texture}. 
Although the scores of PSNR and SSIM are similar, our qualitative evaluation (Figure~\ref{fig:blending_texture}) shows improvements in maintaining texture details present in input images, which is reflected in the LPIPS score (Table~\ref{table:blending_texture}).
The model trained without blending fails to produce clear details of the logo or patterns present in the source image as indicated in the orange box.

\begin{table}[ht]
\begin{center}
\caption{Ablation study for the model trained with and without the texture blending.}
    \begin{tabular}{l|ccc}
    \noalign{\smallskip}
    \hline
    Method
    & LPIPS↓ & PSNR↑          & SSIM↑ \\
    \hline
    \hline
    w/o texture blending
    & 0.2276          & \textbf{18.11} & \textbf{0.6007} \\
    w/ texture blending
    & \textbf{0.2144} & 18.06          & 0.5965 \\
    \hline
    \noalign{\smallskip}
    \end{tabular} \label{table:blending_texture}
\end{center}
\end{table}

\section{Discussion}

\subsection{Limitations}
Even though our method achieves better results compared to previous methods, we have limitations that need to be addressed. Our method generates textures based on the surface of the target 3D human mesh which is predicted by Tex2Shape~\cite{alldieck2019tex2shape}. Because Tex2Shape deforms the SMPL~\cite{loper2015smpl} model to obtain the 3D human mesh, our model cannot fully handle the textures with loose clothes, such as skirts and long coats.

Similar to other approaches based on deep neural networks, the capability of our method is limited to the training dataset. The size of our training dataset is relatively small compared to previous studies. Thus, our method fails to generate a texture map with various garments such as hats or eye glasses as shown in Figure~\ref{fig:qualitative_texture}(h) and the last column in Figure~\ref{fig:SHHQ}. Another example is human identity. The SURREAL dataset is limited to a single face-identity, and our newly collected dataset consists of limited ethnicity. This hinders the model from generating a texture map with the personal identity because the same face appeared in the source image. With a bigger and more diverse dataset, this problem will be alleviated.

\subsection{Future work}
Our method is based on a supervised setting, which requires ground truth data for training. An interesting future research direction is to utilize high-quality image or video datasets, which are relatively easier to acquire, for unsupervised or self-supervised training as exemplified by Grigorev et al.~\cite{grigorev2021stylepeople}. 
Another interesting direction is to exploit the high-capability of GAN models~\cite{karras2019style,fu2022styleganhuman} to generate different views of the source image to achieve better initialization as attempted in video-based methods \cite{alldieck2018detailed, alldieck2018video, bhatnagar2019multi, zhi2020texmesh, mir2020learning, bhatnagar2019multi}.

\section{Conclusion}
In this work, we proposed a method for generating a 3D human texture map from a single image. The key idea of our approach is to complete the incomplete partial texture map by using a sampling network followed by adjusting the resulting texture with a refiner network. Compared to previous approaches, our method generates a texture map with improved quality by successfully retaining the textural patterns presented in the visible regions of the source image while maintaining spatial alignment with the surface of the target 3D human mesh. In addition, our method produces a plausible texture map from a non-frontal view image.
We verified the effectiveness of curriculum learning and other design choices we made by ablation studies and showed significant quality improvement in the resulting texture by comparing our approach with previous methods. 

\section*{Acknowledgements}
We thank the anonymous reviewers for their invaluable comments; Gyeong Hun Im for collecting the various pose data; Jaedong Kim for his assistance in implementing the mapping function. This work was supported by Institute of Information \& communications Technology Planning \& Evaluation (IITP) grant funded by the Korea government (MSIT) (No.2020-0-00450, A Deep Learning Based Immersive AR Content Creation Platform for Generating Interactive, Context and Geometry Aware Movement from a Single Image)

{\small
\bibliographystyle{ieee_fullname}
\bibliography{egbib}
}

\end{document}